\documentclass[letterpaper]{article} 
\usepackage{aaai24}  
\usepackage{times}  
\usepackage{helvet}  
\usepackage{courier}  
\usepackage[hyphens]{url}  
\usepackage{graphicx} 
\usepackage{natbib}  
\usepackage{caption} 

\usepackage{amsmath}
\usepackage{amssymb}
\usepackage{xcolor}
\usepackage{booktabs}

\usepackage{algorithm}
\usepackage{algorithmic}

\usepackage{newfloat}
\usepackage{listings}
\DeclareCaptionStyle{ruled}{labelfont=normalfont,labelsep=colon,strut=off} 
\floatstyle{ruled}
\newfloat{listing}{tb}{lst}{}
\floatname{listing}{Listing}
\title{Multi-Modal Discussion Transformer: Integrating Text, Images and Graph Transformers to Detect Hate Speech on Social Media}
\author {
    Liam Hebert,
    Gaurav Sahu,
    Yuxuan Guo,
    Nanda Kishore Sreenivas, \\
    Lukasz Golab, 
    Robin Cohen
}
\affiliations {
    University of Waterloo \\
    liam.hebert@uwaterloo.ca
}

\usepackage{bibentry}

\begin{document}

\maketitle

\begin{abstract}
We present the Multi-Modal Discussion Transformer (mDT), a novel method for detecting hate speech on online social networks such as Reddit discussions. In contrast to traditional comment-only methods, our approach to labelling a comment as hate speech involves a holistic analysis of text and images grounded in the discussion context. This is done by leveraging graph transformers to capture the contextual relationships in the discussion surrounding a comment and grounding the interwoven fusion layers that combine text and image embeddings instead of processing modalities separately. To evaluate our work, we present a new dataset, HatefulDiscussions, comprising complete multi-modal discussions from multiple online communities on Reddit. We compare the performance of our model to baselines that only process individual comments and conduct extensive ablation studies. 

\end{abstract}
\section{Introduction}
Social media have democratized public discourse, enabling users worldwide to freely express their opinions and thoughts. As of 2023, the social media giant Meta has reached 3 billion daily active users across its platforms \cite{2023metaq1earnings}. While this level of connectivity and access to information is undeniably beneficial, it has also resulted in the alarming rise of hate speech 
\cite{das2020hate}. This pervasive spread of hateful rhetoric has caused significant emotional harm to its targets \cite{janikke2019hate}, triggered social divisions and polarization \cite{Waller2021}, and has caused substantial harm to the mental health of users \cite{wachs2022online}. 
There is an urgent need for a comprehensive solution to automate identifying hate speech as a critical first step toward combatting this alarming practice.

Initially, automated hate speech detection models were limited to text-only approaches such as HateXplain~\cite{mathew2021hatexplain}, which classify the text of individual comments. Such methods have two significant weaknesses. First, social media comments have evolved to include images, which can influence the context of the accompanying text. For instance, a comment may be innocuous, but including an image may transform it into a hateful remark.  Second, hate speech is contextual.  Social media comments are often conversational and are influenced by other comments within the discussion thread. For example, a seemingly innocuous comment such as ``That's gross!" can become hateful in a discussion about immigration or minority issues. 

Ongoing research to address these weaknesses includes multi-modal transformers such as VilT~\cite{kim_vilt_2021} that combine images and text for a richer representation of comments. Still, they do not account for the contextual nature of hate speech. \citet{hebert2022predicting} model discussion context with graph neural networks, but they do not discuss how to integrate the interpretation of images within hateful social media discussions. 
Furthermore, the sequential nature of the proposed architecture prevents text embeddings from being grounded to other comments in a graph. The initial semantic content encoded by a comment embedding may differ when considered with different sets of comments versus in isolation. 

To overcome the limitations of existing methods, we propose the Multi-Modal Discussion Transformer (mDT) to holistically encode comments with multi-model discussion context for hate speech detection. To evaluate our work, we also present a novel dataset, HatefulDiscussions, containing complete multi-modal discussion graphs from various Reddit communities and a diverse range of hateful behaviour. 

We compare mDT against comment-only and graph methods and conduct an ablation study on the various components of our architecture. We then conclude by discussing the potential for our model to deliver social value in online contexts by effectively identifying and combating anti-social behaviour in online communities. We also propose future work towards more advanced multi-modal solutions that can better capture the nuanced nature of online behaviour.

To summarize our contributions: \textbf{1)} We propose a novel fusion mechanism as the core of mDT that interweaves multi-modal fusion layers with graph transformer layers, allowing for multi-modal comment representations actively grounded in the discussion context. \textbf{2)} We propose a novel graph structure encoding specific to the conversational structure of social media discussions. \textbf{3)} We introduce a dataset of 8266 annotated discussions, totalling 18359 labelled comments, with complete discussion trees and images to evaluate the effectiveness of mDT. Our work focuses on Reddit, which consists of branching tree discussions. Our codebase, datasets and further supplemental can be found at github.com/liamhebert/MultiModalDiscussionTransformer.

\begin{figure}
    \centering
    \includegraphics[width=\linewidth]{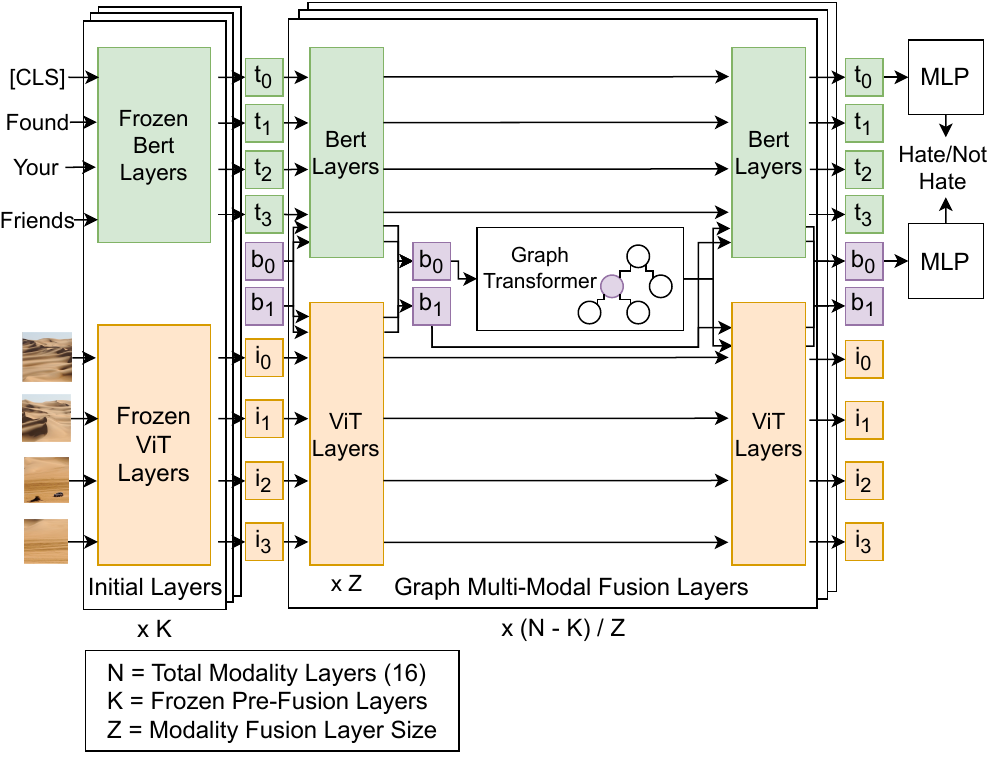}
    \caption{Multi-Modal Discussion Transformer}
    \label{fig:mdt-arch}
\end{figure}

\section{Related Work}
Transformer-based encoding models such as BERT have significantly improved natural language processing due to their ability to capture textual semantics \cite{devlin2018bert}. Inspired by these developments, methods such as HateXplain \cite{mathew2021hatexplain} and HateBERT \cite{caselli-etal-2021-hatebert} have been introduced to discern hateful comments on social platforms, focusing on text alone. The effectiveness of these efforts is intrinsically tied to the diversity of datasets they are trained on. For instance, HateXplain utilized a specialized dataset from diverse social platforms like Twitter and Gab, emphasizing interpretable hate speech detection. Other noteworthy datasets include \citet{gong2021abusive}, studying heterogeneous hate speech (comments containing mixed abusive and non-abusive language), and \citet{founta2018large}, which crowdsourced annotation of Twitter abusive content. Finally, \citet{zampieri-etal-2019-predicting} collected hateful Twitter posts collated through a collaborative semi-supervised approach.

While text is essential, images also contribute to the semantic context. CLIP introduced an approach to align text and image representations via contrastive pre-training \citep{radford2021learning}. ViLBERT \cite{lu_vilbert_2019} conceptualized distinct transformers for each modality—images and text, which are then amalgamated through co-attentional transformer layers. Subsequent works such as VilT \citep{kim_vilt_2021} and \citet{nagrani_bottleneck_2021} have devised novel inter-modality fusion mechanisms, unifying both modality transformers into one. This integration of multi-modal language grounding has also enriched hate speech detection, as evidenced by the HatefulMemes challenge \citep{kiela_hateful_2021}. Additional works such as \citet{liang-etal-2022-multi} employ graph convolutional networks to merge text and images, primarily for sarcasm detection. Meanwhile, \citet{sahu2021towards} leverage generative adversarial networks to encode these modalities, facilitating congruent representations of comments. \citet{cao-etal-2022-prompting} pursue a unique strategy by mapping the paired image to text descriptors, appending the comment text, and then predicting with a generative language model. Finally, \cite{singh2022sentiment} incorporate image and text representations of product reviews to accurately disambiguate complaints. 

Despite the progress, many of these techniques overlook a vital modality: the context of discussions. The prevailing emphasis remains on datasets and techniques that analyze singular comments, bypassing the contextual significance of the prior discussion. By extending Graphormer \citep{ying_transformers_2022}—a graph transformer network tailored for molecular modelling—\citet{hebert2022predicting} consolidate learned comment representations to predict the trajectory of hateful discussions. However, this work has limitations; it neglects the influence of images and, owing to the absence of complete discussion-focused hate speech datasets, resorts to approximating ground truth labels using a ready-made external classifier. Our work addresses both limitations, including interleaving comment and discussion layers and human ground truth data.

\section{Methodology} 
\subsection{Multi-Modal Discussion Transformer (mDT)}

The mDT architecture consists of Initial Pre-Fusion, Modality Fusion, and Graph Transformer (Figure \ref{fig:mdt-arch}). The description below outlines the holistic nature of our solution.

\subsubsection{Initial Pre-Fusion}

Given a discussion $D$ with comments $c \in D$, each represented with text $t_c$ and optional image $i_c$, we start with pre-trained BERT and ViT models to encode text and images, respectively. Both models contain $N$ layers with the same hidden dimension of $d$. In our experiments, we utilized BERT-base and ViT-base, which have $N = 16$ layers and $d = 768$ hidden dimensions. Given these models, the Initial Pre-Fusion step consists of the first $K$ layers of both models with gradients disabled (frozen), denoted as
\begin{align}
   t^k_c = BERT_{init}(t_c), i^k_c = ViT_{init}(i_c) 
\end{align}
where $K < N$. This step encodes a foundational understanding of the images and text that make up each comment.

\subsubsection{Modality Fusion}
After creating initial text and image embeddings $t_c, i_c$ for all comments $c \in D$ in the discussion, we move to the Modality Fusion step. We adopt the bottleneck mechanism proposed by \citet{nagrani_bottleneck_2021} to encode inter-modality information. We concatenate $b$ shared modality bottleneck tokens $B \in R_{b \times d}$ to $t_c$ and $i_c$, transforming the input sequence to $[t^k_c~||~B], [i^k_c~||~ B]$. We then define a modality fusion layer $l$ as 
\begin{align}
[t_c^{l+1} || B_{t, c}^{l+1}] &= BERT_l([t_c^{l} || B_c^{l}]) \\
[i_c^{l+1} || B_{i, c}^{l+1}] &= ViT_l([i_c^{l} || B_c^{l}]) \\
B_c^{l + 1} &= Avg(B_{t, c}^{l + 1}, B_{i, c}^{l + 1})
\end{align} 
where both modalities only share information through the $B$ bottleneck tokens. This design forces both modalities to compress information to a limited set of tokens, improving performance and efficiency. If no images are attached to a comment, then $B_c^{l + 1} = B_t^{l + 1}$. 

\subsubsection{Graph Transformer}
Then, after $Z$ $(< (N - K))$ modality fusion layers, we deploy Graph Transformer layers to aggregate contextual information from the other comments in the discussion\footnote{Our implementation can handle discussions up to 516 comments, as we only require a single graph transformer pass to evaluate all comments. The above limit can be exceeded via efficient attention mechanisms such as sparse or flash attention.}. Given that the tokens in $B_c$ encode rich inter-modality information, we innovate by leveraging these representations to represent the nodes in our discussion graph.  
Using $b^0_c \in B_c$ to represent each comment $c \in D$, we aggregate each embedding using a transformer model to incorporate discussion context from other comments. Our novel utilization of bottleneck tokens to represent graph nodes allows modality models to maintain a modality-specific pooler token ([CLS]) as well as a graph context representation ($b_0$). 

Since transformer layers are position-independent, we include two learned structure encodings. The first is Centrality Encoding, denoted $z$, which encodes the degree of nodes in the graph \cite{ying_transformers_2022}. Since social media discussion graphs are directed, the degree of comments is equivalent to the number of replies a comment receives plus one for the parent node. We implement this mechanism as
\begin{align}
    h_c^{(0)} = b_c^0 + z_{deg(c)}
\end{align}
where $h_c^{(0)}$ is the initial embedding of $b_c^0$ in the graph and $z_{deg(c)}$ is a learned embedding corresponding to the degree $deg(c)$ of the comment.   

The second structure encoding is Spatial Encoding, denoted $s_{(c, v)}$, which encodes the graph's structural relationship between two nodes, $c$ and $v$. This encoding is added as an attention bias term during the self-attention mechanism. That is, we compute the self attention $A_{(c, v)}$ between nodes $c, v$ as
\begin{align}
  A_{(c,v)} = \frac{(h_c \times W_Q)(h_v \times W_K)}{\sqrt{d}} + s_{(c,v)}  
\end{align}
where $W_Q$ and $W_K$ are learned weight matrices and $d$ is the hidden dimension of $h$. 

\begin{figure}
    \centering
    \includegraphics{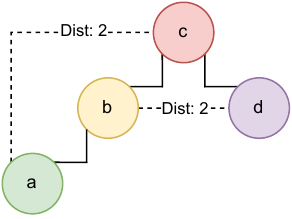}
    \caption{Example Discussion Structure. Each node in the discussion tree represents a comment. The shortest distance between (a, c) and (b, d) is equivalent, demonstrating a lack of expressiveness towards hierarchy.}
    \label{fig:sample_structure}
\end{figure}

In previous graph transformer networks, $s_{(c, v)}$ is encoded as a learned embedding representing the shortest distance between nodes $c$ and $v$ in the graph \cite{ying_transformers_2022, hebert2022predicting}. However, this metric does not lend itself well to the hierarchical structure of discussions, where equivalent distances can represent different interactions. This is best seen in the example discussion illustrated in Figure \ref{fig:sample_structure}.
When utilizing the shortest distance to encode structure, the distance between nodes $a$ and $c$ is the same as the distance between nodes $b$ and $d$ in this graph. However, $b$ and $d$ represent direct replies to the same parent post, whereas $a$ is two comments underneath $c$.

We propose a novel hierarchical spatial encoding based on Cantor’s pairing function to account for this. Cantor's pairing function uniquely maps sets of two numbers into a single number $\mathbb{N} \times \mathbb{N} \rightarrow \mathbb{N}$. 
Given comments $a$ and $b$, we first calculate the number of hops upward $u_{(a, b)}$ and hops downward $d_{(a, b)}$ to reach $b$ from $a$. In the example above, the distance between $a$ and $d$ is $u_{(a, b)} = 2, d_{(a, b)} = 1$. We then compress both numbers into a single index using the proposed position-independent variant of Cantor’s pairing:
\begin{align}
s_{(c, v)} &= s_{(v,c)} \\
&= Cantors(u, d)\\
&= \frac{(u + d)(u + d + 1)}{2} + min(u, d) 
\end{align}
This uniquely maps $\mathbb{N} \times \mathbb{N} \rightarrow \mathbb{N}$ such that $s_{c,v} = s_{v,c}$. We utilize this function to index learned spatial embeddings in the self-attention mechanism. 

After $G$ graph transformer layers, the final representation of $h_c^{G}$ replaces $b_c^0$ for the next set of $Z$ modality fusion layers. We denote the combination of $Z$ Modality Fusion and $G$ Graph Transformer layers as a Graph Multi-Modal Fusion module. Finally, after $(N - K) / Z$ Graph Multi-Modal Fusion modules, we predict logits using the final embedding of $b_c^0$ and the [CLS] embedding of $t_c$. This novel interweaving of graph transformer and fusion layers through modality bottleneck tokens ensures that fusion models create representations grounded in the discussion context. Notably, this differs from previous approaches that utilize graph neural networks, which sequentially process individual comments before applying a set of graph layers. 

\subsection{HatefulDiscussions Dataset}
To train our model, we require a complete multi-modal discussion graph dataset. 
However, datasets used by other works \cite{mathew2021hatexplain, zampieri-etal-2019-predicting, kiela_hateful_2021} consist of individual labelled comments and are predominately text-only. 
To address this issue, we curated a novel benchmark comprising multiple datasets that used human annotators, which we augmented to include complete multi-modal discussion graphs. Our final dataset comprises 8266 Reddit discussions with 18359 labelled comments from 850 communities.
Note that our architecture can extend to other platforms, such as Facebook, Twitter and TikTok, as they also discuss in tree structures. Since discussions on these platforms are typically smaller with less complexity, using Reddit allows us to best stress-test our model. 

The first type of hate speech included in our benchmark is Identity-Directed and Affiliation-Directed Abuse. We retrieved labelled examples of this from the Contextual Abuse Dataset (CAD) developed by \citet{vidgen-etal-2021-introducing}.
According to the authors, Identity-Directed abuse refers to content containing negative statements against a social category, encompassing fundamental aspects of individuals' community and socio-demographics, such as religion, race, ethnicity, and sexuality, among others. On the other hand, Affiliation-Directed abuse is defined as content expressing negativity toward an affiliation, which is described as a voluntary association with a collective, such as political affiliation and occupations \cite{vidgen-etal-2021-introducing}. We selected both of these forms of abuse from CAD due to the similarity in their definitions—abuse that is directed at aspects of a person's identity rather than a specific individual directly.

Next, slurs form the second type of hateful content within our dataset, sampled from the Slurs corpus \cite{kurrek-etal-2020-towards}. Notably, historically derogatory slurs can undergo re-appropriation by specific communities, such as the n-slur in African American Vernacular, transforming them into non-derogatory terms. Therefore, we hypothesize that understanding the contextual nuances surrounding the use of slurs becomes essential in distinguishing between non-derogatory and derogatory instances. 

The last type of hateful content we include is person-directed abuse, hate speech or offensive content that specifically targets and attacks an individual or a group of individuals. We source labelled examples from the Learning to Intervene (LTI) dataset by \citet{qian-etal-2019-benchmark} to include examples of this abuse requiring context.

\begin{table}
    \centering
    \begin{tabular}{ll}
    \toprule
        Label & Count \\
        \midrule
        Derogatory Slur & 4297 \\
        Not Derogatory Slur (NDG) & 2401 \\
        Homonym (HOM) & 364 \\
        \midrule 
        LTI Person Directed Neutral & 4116 \\ 
        LTI Person Directed Hate & 1313 \\ 
        \midrule 
        CAD Affiliation Neutral & 4892 \\ 
        CAD Identity Directed Hate & 701 \\
        CAD Affiliation Directed Hate & 275 \\
        \midrule
        Neutral & 11773 \\ 
        Hateful & 6586 \\
        \bottomrule
    \end{tabular}
    \caption{Label Distribution of Hateful Discussions}
    \label{tab:dist_labels}
\end{table}

For each labelled comment, we retrieved the corresponding complete discussion tree using the Pushshift Reddit API and downloaded all associated images\footnote{At the time of writing, Reddit has suspended access to the Pushshift API; however, our dataset remains complete.}. To refine our dataset, we filtered out conversations without images and constrained comments to have a maximum degree of three and conversations to have a maximum depth of five. By trimming the size of the discussion tree, we reduce computational complexity and focus on the most relevant parts of the conversation \cite{parmentier2021}. We map each retrieved label to Hateful or Normal and treat the problem as a binary classification. The distribution of each label can be seen in Table \ref{tab:dist_labels}. 

Most images in HatefulDiscussions, such as the root post, appear in the discussion context rather than directly attached to labelled comments. In our case, only 424 labelled instances have an image attached. Still, all 8000 discussions have an image in the prior context. Therefore, the challenge becomes how to interpret and incorporate multi-modal discussion context to disambiguate the meaning of comments that may not contain those modalities.    

\section{Results}

\subsection{Experimental Setup}
\begin{table}
    \centering
    \begin{tabular}{ll}
        \toprule
        Hyperparameter & Value \\
        \midrule\midrule
        Pre-Fusion Layers (K) & 4 - [0, 2, 4, 6]  \\
        Modality Fusion Layer Stack (Z) & 2 (8 total) - [1, 2, 4] \\
        Graph Layer Stack (G) & 2 (8 total) - [1, 2, 4] \\
        Bottleneck Size (B) & 4 - [1, 4, 8, 16, 32] \\
        Max Spatial Attention & 5 - [2, 5, 4096 (inf)] \\
        \midrule
        Attention Dropout & 0.3 - [0, 0.1, 0.3, 0.5] \\
        Activation Dropout & 0.3 - [0, 0.1, 0.3, 0.5] \\ 
        Graph Dropout & 0.4 - [0, 0.2, 0.4, 0.6] \\
        \midrule
        Hidden Dimension (d) & 768 \\
        Graph Attention Heads & 12 - [4, 6, 12, 24] \\
        Modality Attention Heads & 12 \\
        \midrule
        Optimizer & Adam \\ 
        Batch Size & 48 \\ 
        Epochs & 10 w/ Early Stopping \\
        \midrule
        Learning Rate & $3e^{-5} \rightarrow 3e^{-7}$ \\
        Learning Rate Scheduler & Polynomial Decay \\ \midrule
        Warm up Updates & 500 \\
        Total Updates & 3350 \\
        \midrule
        Positive Class Weighting & 1.5 - [1, 1.5] \\
        Negative Class Weight & 1 \\
        Freeze Initial Encoders & Yes - [Yes, No] \\
        \bottomrule
    \end{tabular}
    \caption{mDT Model Hyperparameters. The search space for each parameter is denoted by [...]}
    \label{tab:hyperparameters}
\end{table}
We conduct a 7-fold stratified cross-validation with a fixed seed (1) and report the average performance for each model. We report overall accuracy (Acc.) and class-weighted Precision (Pre.), Recall (Rec.) and F1 to account for label imbalance. Underlined values denote statistical significance using Student's t-test with p-value $< 0.05$ in all results. Our hyperparameter search space can be seen in Table \ref{tab:hyperparameters}. All experiments were run using 2xA100-40GB GPUs, a 12-core Intel CPU, 80GB of RAM, and Linux. 

\subsection{Text-only Methods vs. Discussion Transformers}
\begin{table}
    \centering
    \begin{tabular}{lcccc}
    \toprule
        Method & Acc. & Pre. & Rec. & F1 \\
        \midrule
        BERT-HateXplain & 0.742 & 0.763 & 0.742 & 0.747 \\
        Detoxify & 0.687 & 0.679 & 0.696 & 0.677\\ 
        RoBertA Dynabench & 0.811 & 0.822 & 0.811 & 0.814 \\
        BERT-HatefulDiscuss & 0.858 & 0.858 & 0.858 & 0.858  \\
        Graphormer & 0.735 & 0.594 & 0.759 & 0.667 \\
        mDT (ours) & \textbf{\underline{0.880}} & \textbf{\underline{0.880}} & \textbf{\underline{0.880}} & \textbf{\underline{0.877}} \\
        \bottomrule
    \end{tabular}
    \caption{Performance of mDT against Text-Only Methods}
    \label{tab:textvsgraph}
\end{table}

To assess the performance of mDT, we compared it against several state-of-the-art hate speech detection methods. For comment-only approaches, we evaluated BERT-HateXplain \cite{mathew2021hatexplain}, Detoxify \cite{Detoxify}, and RoBertA Dynabench \cite{vidgen2021lftw}. We also compared mDT against a BERT model trained on the training set of HatefulDiscussions, referred to as BERT-HatefulDiscuss. To compare against previous graph-based approaches, we evaluated the text-only Graphormer model proposed by \cite{hebert2022predicting}.

Our results (Table~\ref{tab:textvsgraph}) show that mDT outperforms all evaluated methods across all metrics. Specifically, mDT achieves 14.5\% higher accuracy and 21\% higher F1 score than Graphormer. This indicates that our approach to including graph context significantly improved over the previous approach incorporating this modality. Although the performance gap between BERT-HatefulDiscussions and mDT is narrower, we still perform better against all text-only methods. We observed F1 score improvements of 20\%, 13\%, and 6.3\% over Detoxify, BERT-HateXplain, and RoBertA Dynabench, respectively.

\subsection{Effect of Bottleneck Size}
\begin{table}
    \centering
        
    \begin{tabular}{ccccc}
    \toprule
        Bottleneck Size & Acc. & Pre. & Rec. & F1  \\
        \midrule
        4 & \textbf{0.880} & \textbf{\underline{0.880}} & \textbf{0.880} & \textbf{0.877} \\
        8 & 0.863 & 0.864 & 0.863 & 0.863\\ 
        16 & 0.864 & 0.850 & 0.853 & 0.852\\ 
        32 & 0.874 & 0.872 & 0.874 & 0.872\\
        \bottomrule
    \end{tabular}
    \caption{Effect of Bottleneck Size on mDT Performance}
    \label{tab:bottleneck}
\end{table}
Next, we investigated the impact of increasing the number of bottleneck interaction tokens ($B$) in mDT, added during the modality fusion step. Adding more bottleneck tokens reduces the amount of compression required by the BERT and ViT models to exchange information. Table \ref{tab:bottleneck} presents the results, where we find that using four bottleneck tokens leads to the best performance. We also observe a slight drop in performance when we increase the number of bottleneck tokens beyond four, indicating the importance of compression when exchanging modality encodings between models. 

\subsection{Effect of Constrained Graph Attention}
\begin{table}
    \centering
    \begin{tabular}{ccccc}
    \toprule
        Attention Window & Acc. & Pre. & Rec. & F1  \\
        \midrule
        2 & 0.866 & 0.866 & 0.866 & 0.866\\
        5 & \textbf{\underline{0.880}} & \textbf{\underline{0.880}} & \textbf{\underline{0.880}} & \textbf{\underline{0.877}} \\
        $\infty$ & 0.870 & 0.861 & 0.850 & 0.855\\
        \bottomrule
    \end{tabular}
    \caption{Effect of Constraining Graph Attention}
    \label{tab:constrained_attention}
\end{table}
A recent study by Hebert et al. explored the limitations of graph transformers for hate speech prediction, finding that discussion context can sometimes mislead graph models into making incorrect predictions \cite{hebert2023qualitative}. In light of this, we explore the impact of constraining the attention mechanism of our graph transformer network to only attend to nodes within a maximum number of hops away from a source node. We report the results in Table \ref{tab:constrained_attention} and find that constraining the attention window to 5 hops achieves better performance. However, we also observed that performance gains from the 5-hop constraint were lost when we further constrained the attention to only two hops. Our findings suggest a balance is required when constraining graph attention for optimal performance.

\subsection{Effect of Fusion Layers}
\begin{table}
    \centering
    \begin{tabular}{ccccc}
    \toprule
        Fusion Layers & Acc. & Pre. & Rec. & F1  \\
        \midrule
        6 & 0.868 & 0.856 & 0.854 & 0.855\\
        8 & 0.872 & 0.871 & 0.844 & 0.855 \\ 
        10 & 0.866 & 0.867 & 0.866 & 0.862\\ 
        12 & \textbf{\underline{0.880}} & \textbf{\underline{0.880}} & \textbf{\underline{0.880}} & \textbf{\underline{0.877}} \\
        \bottomrule
    \end{tabular}    
    \caption{Effect of Fusion Layers}
    \label{tab:fusion_layers}
\end{table}
Next, we investigate the effect of increasing the number of Multi-Modal Fusion Layers ($Z$) in our mDT model. To ensure full utilization of the 16 available layers, any unused layers were allocated to the Initial Pre-Fusion step ($K$). Our results in Table \ref{tab:fusion_layers} indicate that utilizing 12 fusion layers leads to the best performance. Interestingly, the performance gains did not follow a linear trend with the number of fusion layers. Specifically, we observed that eight fusion layers outperformed ten but were still inferior to 12. Further research should explore the potential benefits of scaling beyond 12 fusion layers using larger modality models.  

\subsection{Effect of Images}
\begin{table}
    \centering
    \begin{tabular}{lcccc}
    \toprule
        Usage of Images & Acc. & Pre. & Rec. & F1  \\
        \midrule
        With Images  & \textbf{\underline{0.880}} & \textbf{\underline{0.880}} & \textbf{\underline{0.880}} & \textbf{\underline{0.877}} \\
        Without Images & 0.832 & 0.835 & 0.822 & 0.828 \\
        \bottomrule
    \end{tabular}
    \caption{Effect of Excluding Images}
    \label{tab:images}
\end{table}
We also investigated the impact of removing images in mDT. Our findings (Table \ref{tab:images}) support the hypothesis that images provide crucial contextual information for detecting hateful content: excluding images led to a 4.8\% decrease in accuracy and a 4.9\% decrease in the F1 score. It is worth noting that even without images, mDT outperformed Graphormer (Table \ref{tab:textvsgraph}), indicating that our approach provides substantial gains over previous graph-based methods for hate speech detection beyond just including images. The results of this experiment underscore the importance of considering multiple modalities for hate speech detection and suggest that future research should explore further improvements by leveraging additional types of contextual information.

\subsection{Qualitative Analysis: BERT vs.\ mDT}
\label{sec:qual_analysis}

\begin{figure}
    \centering
    \includegraphics[width=0.85\linewidth]{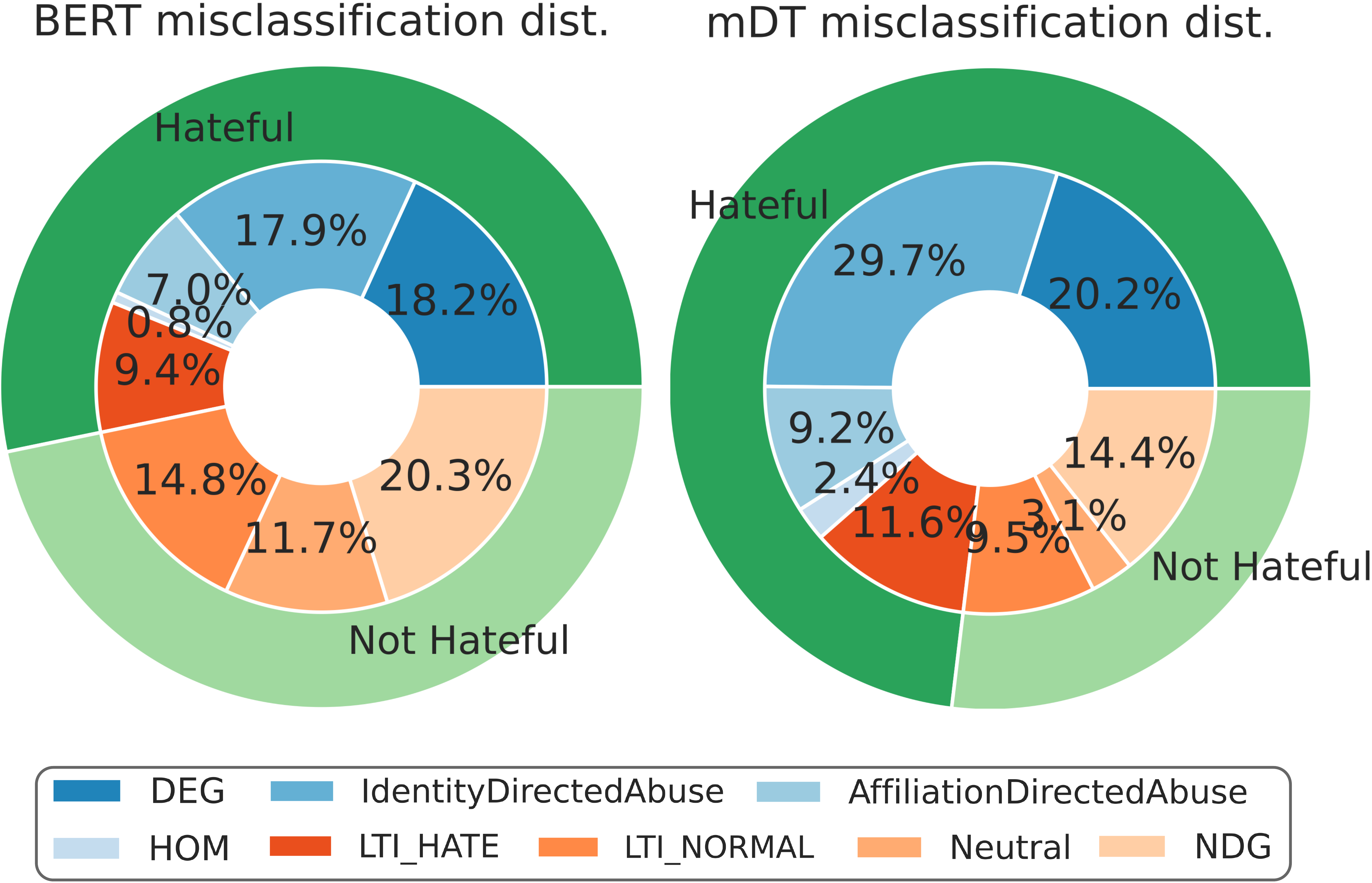}
    \caption{Fine-grained distribution of BERT and mDT misclassification. (Acronyms above as in Table \ref{tab:dist_labels})}
    \label{fig:failure_cases}
\end{figure}

\begin{table*}[t]
    \small
    \centering
    \resizebox{0.9\textwidth}{!} {
    \begin{tabular}{|p{5cm}|p{5cm}|c|c|}
        \hline
        \textbf{Primary Text} & \textbf{Context (only seen by mDT)} & BERT pred. & mDT pred. \\
        \hline
        Now imagine if virtuous keyboard sjws had their way? Their mascot should be Ralph Wiggum. & \textbf{[\dots]} Preferred pronouns: go/\textit{f-slur}/yourself \textbf{[\dots]} If the Chinese in my corner of NZ only sold to Chinese they'd starve by Thursday. \textbf{[\dots]} They just wanna \textit{b-slur} about something because their own life sucks. & Not Hateful & Hateful \\
        \hline
        whoa brah.. leave my tranny out of this & [image] (Figure \ref{fig:meme}) \textbf{[\dots]} that's \textit{f-slur} retarded \textbf{[\dots]} Just spit my drink \textbf{[\dots]} & Not Hateful & Hateful \\
        \hline
        "That \textit{n-slur} was on PCP Johnson" Lmao & [image] (Figure \ref{fig:pcp}) \textbf{[\dots]} Its' a common pattern when dealing with these shootings. * Kill black dude \textbf{[\dots]} * Wingnut welfare kicks in as racist \textit{f-slur} create gofundme of over half a million \textit{f-slur} dollars for cops family \textbf{[\dots]} & Not Hateful & Hateful\\
        \hline
        uwu owo uwu & \textbf{[\dots]} That is not even close to what feminism is. What you are talking about is radical Feminism \textbf{[\dots]} Got banned from my sexual minority subreddit (r/bisexual) for not believing that all bisexuals should actually be pansexuals \textbf{[\dots]} & Hateful & Not Hateful \\
        \hline
    \end{tabular}%
    }
    \caption{Text instances misclassified by BERT and mDT. Note: The ground truth for all the examples shown here is ``Hateful". We have also redacted chunks of text from the context in the interest of space. The redacted content is shown by [\dots].} 
    
\label{tab:BERT_vs_graph_examples}
\end{table*}

We next perform a qualitative comparison of the text-only BERT model and the proposed mDT architecture.
We find that the text-only BERT model misclassifies 385/2717 test instances.
Upon passing those test instances through mDT, we found that it corrected BERT's labels in 161/385 cases.
We further note that BERT and mDT predictions disagree on 264 test instances, which mDT is correct on 161 (61\%).
Figure~\ref{fig:failure_cases} shows a fine-grained distribution of misclassified test examples by class.
Using mDT results in an overall decrease in misclassifications (385 $\rightarrow$ 327), with a significant reduction in false positives (fewer misclassifications for the `Not Hateful' class).
However, we notice that BERT and mDT both struggle to detect the presence of hate speech in derogatory slur (DEG) and identity-directed (IdentityDirectedAbuse) comments.

\begin{figure}
    \centering
    \includegraphics[width=0.6\linewidth]{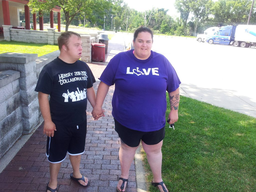}
    \caption{An image present in the discussion context of example 3 (Table \ref{tab:BERT_vs_graph_examples}), seen only by mDT, contextualizing comments as potentially hateful}
    \label{fig:meme}
\end{figure}

\begin{figure}
    \centering
    \includegraphics[width=0.6\linewidth]{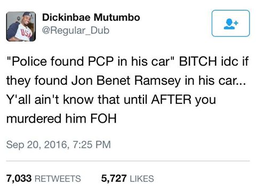}
    \caption{An image present in discussion context of example 2 (Table \ref{tab:BERT_vs_graph_examples}), seen only by mDT, contextualizing comments as potentially hateful}
    \label{fig:pcp}
\end{figure}

Table~\ref{tab:BERT_vs_graph_examples} shows some hateful test instances misclassified by the two models.
We note that the main text under consideration (an individual comment) may not exhibit hate speech on its own; however, considering it with the context (rest of the discussion thread+image) helps mDT correctly classify the test instances as hate speech. Consider the second example in Table~\ref{tab:BERT_vs_graph_examples}. The word ``tranny" is a common acronym for ``transmission" on social media, but considering the context, it is an abusive discussion directed toward the transgender community. This is further contextualized by the accompanying image in the discussion, leading to evidence of hateful interpretations (Figure \ref{fig:meme}). Images within the discussions for each example provide similar contextual evidence, such as the third example and Figure \ref{fig:pcp}. 

We also found some intriguing test examples where adding context proved misleading for the model, while BERT confidently classified the main text as hateful. For instance, in the last example in Table~\ref{tab:BERT_vs_graph_examples}, comments in the context are largely non-abusive, misinterpreting the primary text as non-abusive. This suggests that while adding context results in a net decrease in misclassifications, majorly neutral context might also fool the model. This is likely since we emphasize the discussion context when we obtain the final classification by averaging the text embedding logit with the discussion node embedding ($b_c^0$).

\section{Future Work}
While we find mDT to be an effective method for analyzing discussions on social media, we have pointed out how it is challenged when the discussion context contains predominately neutral comments. To address this, we propose using a first-stage text ranker to compute semantic relevance between comments to filter unrelated messages.

We also note that many contextual signals in social media discussions remain untapped beyond text, images, and discussion structure. Incorporating named entity recognition techniques to integrate deeper analysis of real-world knowledge 
would be well-supported by the contextual nature of mDT\cite{mishra2022tweetnerd}. 

Perhaps the most exciting step forward would be to expand our analysis of individual communities toward learning indicators of their propensity for hateful conduct. We are interested in capturing the culture of specific platforms containing diverse communities, including marginalized communities, which exchange unique reclaimed vernacular that should not be misinterpreted as hate. In addition to the example given earlier of the African American community, there are particular usages as well that arise among platforms supporting LGBTQ users. The contextual nature of mDT captured by graph transformers provides much promise for advancing these extensions.

Finally, the versatility of mDT's core mechanisms makes it a promising tool for many applications beyond hate speech detection. This approach could be applied to other domains such as online product reviews \citep{jagdale2019sentiment}, political discourse analysis \citep{hanjia2022understanding}, and popularity analysis \citep{weilong2022title, yunpeng2022efficient}, where understanding the discussion context is critical for accurate interpretation.

\section{Conclusion}
This paper presented mDT: a holistic approach to detecting hate speech in social media. Our model leverages graphs, text and images to reason about entire discussion threads. Core to our approach is the introduction of hierarchical spatial encodings and coupling of text, image, and graph transformers through a novel bottleneck mechanism. 
We also presented a new dataset of complete multi-modal discussions containing a wide spectrum of hateful content, enabling future work into robust graph-based solutions for hate speech detection\footnote{We leave additional comparisons with image-text multi-modal methods to future work to resolve fairness considerations; these competitors require comments to be multi-modal, whereas we can notice relevant discussion responses with images.}.

One significant contribution is demonstrating how discussion-oriented multi-modal analysis can improve the detection of anti-social behaviour online. Compared with several key competitors, our experimental results demonstrate the quantitative improvements stemming from our method. Notably, we see a 21\% improvement in F1 over previous methods to include discussion context, such as \cite{hebert2022predicting}. Furthermore, our initial qualitative analysis demonstrates the valuable impact of our holistic multi-modal approach. 

Beyond enhanced holistic discussion analysis, our work enables a rich understanding of conversational dynamics, enabling community-centric prediction. This is primarily powered by our novel improvements to graph transformers, a method gaining momentum in AI molecular modelling, revealing their potential to capture the relationships in complex multi-modal discussions. We hypothesize that this expressiveness in capturing context can aid in disambiguating false positives, preventing further marginalization of communities, and proactively mitigating hateful behaviours. 

Overall, our approach presents a path forward for addressing the issue of hate speech on social media and encourages the exploration of holistic graph-based multi-modal models to interpret online discussions. We believe our research can help foster healthier and more inclusive environments, improving mental health for individuals online.

\section{Ethics Statement}
Our data collection efforts to create HatefulDiscussions are consistent with Reddit's Terms of Service and with approval. We removed all user-identifying features and metadata from all posts and images to ensure privacy and refrained from contacting users. To best comply with their terms of service, we will be open-sourcing our dataset and pre-trained models under the CC-BY-NC 4.0 license, with the model code under the permissive MIT license. 

\section{Acknowledgements}
The authors thank the Natural Sciences and Engineering Research Council of Canada, the Canada Research Chairs Program and the University of Waterloo Cheriton Scholarship for financial support. We are also grateful to the reviewers for their valued feedback on the paper.

\bibliography{aaai24}

\end{document}